\documentclass[11pt, letterpaper]{article}
\usepackage{header}

\begin{document}

\title{Accelerated Discovery of Cryoprotectant Cocktails via Multi-Objective Bayesian Optimization}

\author[1]{Daniel Emerson}
\author[2]{Nora Gaby-Biegel}
\author[2]{Purva Joshi}
\author[1]{Yoed Rabin}
\author[2]{Rebecca D. Sandlin}
\author[1]{Levent Burak Kara}

\affil[1]{Mechanical Engineering Department, Carnegie Mellon University, Pittsburgh PA USA}
\affil[2]{Center for Engineering in Medicine \& Surgery, Department of Surgery, Massachusetts General Hospital, Harvard Medical School, and Shriners Children's, Boston MA USA}

\date{}
\maketitle
Designing cryoprotectant agent (CPA) cocktails for vitrification is challenging because formulations must be concentrated enough to suppress ice formation yet non-toxic enough to preserve cell viability. This tradeoff creates a large, multi-objective design space in which traditional discovery is slow, often relying on expert intuition or exhaustive experimentation. We present a data-efficient framework that accelerates CPA cocktail design by combining high-throughput screening with an active-learning loop based on multi-objective Bayesian optimization. From an initial set of measured cocktails, we train probabilistic surrogate models to predict concentration and viability and quantify uncertainty across candidate formulations. We then iteratively select the next experiments by prioritizing cocktails expected to improve the Pareto front, maximizing expected Pareto improvement under uncertainty, and update the models as new assay results are collected. Wet-lab validation shows that our approach efficiently discovers cocktails that simultaneously achieve high CPA concentrations and high post-exposure viability. Relative to a naive strategy and a strong baseline, our method improves dominated hypervolume by 9.5\% and 4.5\%, respectively, while reducing the number of experiments needed to reach high-quality solutions. In complementary synthetic studies, it recovers a comparably strong set of Pareto-optimal solutions using only 30\% of the evaluations required by the prior state-of-the-art multi-objective approach, which amounts to saving approximately 10 weeks of experimental time. Because the framework assumes only a suitable assay and defined formulation space, it can be adapted to different CPA libraries, objective definitions, and cell lines to accelerate cryopreservation development.
\section{Introduction}

Cryopreservation by vitrification is a cornerstone technology for long-term storage of cells, tissues, and biological systems, enabling advances across reproductive medicine, regenerative therapies, organ banking, and basic biological research \cite{fahyVitrificationApproachCryopreservation1984,rallIcefreeCryopreservationMouse1985}. Vitrification suppresses ice crystallization by using highly concentrated cryoprotectant agents (CPAs), thereby avoiding the mechanical and osmotic damage associated with ice formation \cite{fahyCryoprotectantToxicityCryoprotectant1990}. Despite its promise, the widespread adoption of vitrification remains constrained by the toxicity of CPAs, which can severely compromise post-exposure cell viability \cite{bestCryoprotectantToxicityFacts2015}. CPA toxicity is highly dependent on concentration, cell type, loading conditions, and interactions between multiple cryoprotectants, making the design of effective CPA formulations a persistent and significant challenge in cryobiology \cite{szurekComparisonAvoidanceToxicity2011,lawsonCytotoxicityEffectsCryoprotectants2011}.

The prevailing strategy for mitigating CPA toxicity is to combine multiple cryoprotectants into so-called CPA cocktails, leveraging synergistic effects that reduce toxicity at a given total concentration \cite{lawsonCytotoxicityEffectsCryoprotectants2011,fahyCryoprotectantToxicityNeutralization2010}. Empirically, uniquely optimized CPA formulations have been developed for specific biological systems, including mammalian embryos, oocytes, organs, and insect models \cite{mazurCryobiologicalPreservationDrosophila1992,sahaNormalCalvesObtained1996,kuwayamaHighlyEfficientVitrification2005}. Experimental approaches such as median lethal dose characterization \cite{nesbittCryoprotectantToxicityHypothermic2021}, kinetic toxicity modeling \cite{warnerMultipleCryoprotectantToxicity2022}, and high-throughput screening assays \cite{warnerRapidQuantificationMulticryoprotectant2021,jaskiewiczValidationHighthroughputScreening2025} have improved our ability to quantify CPA toxicity and explore formulation spaces. However, these methods remain limited in their ability to efficiently navigate the combinatorial explosion of possible CPA cocktails, particularly when multiple objectives such as concentration and viability must be considered simultaneously.

A key limitation of existing approaches is that they rely heavily on either exhaustive experimentation or fixed experimental designs that scale poorly with the dimensionality of the formulation space. Even with automated high-throughput screening, the number of possible CPA combinations grows rapidly with the number of components and allowable concentration increments, rendering naive search strategies impractical \cite{terayamaBlackBoxOptimizationAutomated2021}. Moreover, CPA toxicity mechanisms are only partially understood and vary across cell lines and experimental conditions, limiting the effectiveness of mechanistic or parametric models \cite{fahyCryoprotectantToxicityCryoprotectant1990}. As a result, current methods struggle to balance experimental efficiency with the need to identify Pareto-optimal trade-offs between competing objectives, such as maximizing vitrification-relevant concentration while preserving cell viability.

In this work, we present a data-efficient framework for CPA cocktail optimization that integrates high-throughput screening with iterative, multi-objective Bayesian optimization. Starting from an initial set of experimentally measured CPA cocktails, we train probabilistic surrogate models that capture both predicted performance and uncertainty across the formulation space. These models are embedded within an active learning loop that selects new CPA cocktails to evaluate by maximizing expected improvement of the Pareto front under uncertainty. By explicitly accounting for both concentration and viability as competing objectives, our approach systematically balances exploration and exploitation, enabling rapid discovery of high-quality CPA formulations. Experimental validation using a validated T24 cell-based assay demonstrates that the proposed framework consistently outperforms random sampling and established scalarization-based baselines, achieving superior hypervolume and inverted generational distance metrics while requiring substantially fewer experiments.

Our main contributions are:
\begin{itemize}
    \item a multi-objective, uncertainty-aware optimization framework for CPA cocktail design that couples high-throughput screening with Bayesian optimization,
    \item a batch-aware active learning strategy that efficiently explores large CPA formulation spaces while targeting Pareto-optimal trade-offs between concentration and viability,
    \item a generalizable methodology for accelerating cryoprotectant development that can be adapted to different CPA libraries, objective definitions, and biological systems.
\end{itemize}

\section{Methods}
\begin{figure}[!htb]
    \centering
    \includegraphics[width=\linewidth]{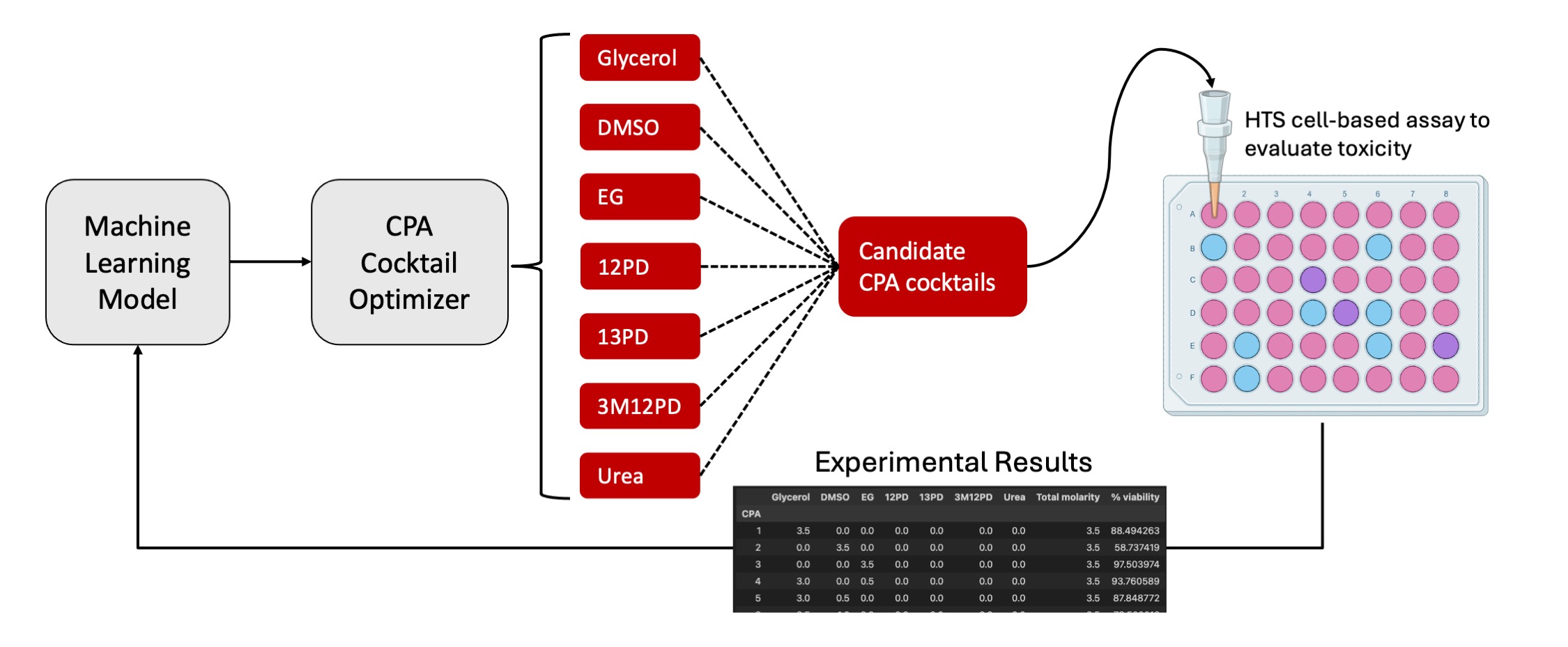}
    \caption{Overview of the iterative CPA cocktail optimization process}
    \label{fig:cpa-opt-overview}
\end{figure}

In this section, we present our methodology to optimize for high-concentration, high viability CPA cocktails in an iterative manner. \cref{fig:cpa-opt-overview} presents an overview of the optimization process, whereby experimental data is obtained through high-throughput screening of candidate CPA cocktails. This data is used to update a machine learning model that predicts the viability of any combination of the component CPAs. An optimization algorithm is used to select the next most informative batch of candidate CPA cocktails, and the process iterates.

\subsection{High Throughput Screening}
We have previously established a high-throughput screening assay to assess the viability of individual CPAs and CPA cocktails \cite{jaskiewiczValidationHighthroughputScreening2025}. After assessing eight cell lines, the T24 cell line was shown to exhibit favorable properties, namely strong plate attachment and resistance to high osmotic gradients. Next, the T24-based assay was validated with positive and negative controls, and the Z-factor was calculated at 0.67, indicating excellent assay quality. Using a BioTek MultiFlo FX automated liquid dispenser, individual CPAs and CPA cocktails can be assessed in 0.5M increments. There are seven component CPAs considered in our study, namely glycerol, dimethyl sulfoxide (DMSO), ethylene glycol (EG), propane-1,2-diol (12PD), propane-1,3-diol (13PD), 3-methyl propane-1,2-diol (3M12PD), and urea. Accounting for on-plate controls, we can test 40 unique CPAs or CPA cocktails with each iteration of HTS.

To generate an initial set of data, we performed LD50 testing of the seven component CPAs for concentrations $C \in [0.5, 5.0]\text{M}$ in 0.5M increments, resulting in 70 data points. Then we generated an additional set of 465 CPA cocktails with concentrations $C \in [3.5, 5.0]\text{M}$. In total, we have a set of 535 data points. To put the size of this dataset into perspective, we are interested in determining high-concentration, high-viability CPA cocktails. If we were to generate all of the possible combinations of cocktails with $C \in [3.5, 6.0]\text{M}$ with component CPAs in 0.5M increments, there would be 48,198 possible cocktails. 

Rather than fitting a defined mathematical model to this data like Warner et al. \cite{warnerMultipleCryoprotectantToxicity2022}, we can fit any number of regression or machine learning models to predict the viability of a candidate CPA cocktail, provided some combination of the seven component CPAs. While not all machine learning models provide an explicit functional form, multilayer perceptrons have been shown to be universal function approximators \cite{hornikMultilayerFeedforwardNetworks1989}. The dilemma we face when fitting a model to this data is ensuring that our model accurately predicts the viability of CPA cocktails that are not seen in the dataset, especially considering our experimental data only covers a small portion of the possible design space.

\subsection{Active Learning}
\label{sec:active-learning}
The dilemma in the previous section is well addressed by ``active learning'' approaches, which consider cases where unlabeled data is abundant, but obtaining labels is an expensive, time-consuming process \cite{settlesActiveLearningLiterature2009}. Let $T$ be the total set of data under consideration. For each iteration $i$ of the active learning algorithm we break $T$ up into the following subsets: data points where the label is known, $T_{K,i}$, data points where the label is unknown $T_{U,i}$, and a subset of $T_{U,i}$ that is chosen to be labeled, $T_{C,i}$. Active learning at its core considers how we determine, $T_{C,i}$, the data points that we need to label.

For our problem, there are several important considerations that rapidly narrow the active learning approaches available to us. First, our problem is considered a regression problem rather than a classification problem; we are trying to determine a numerical value for the viability of each CPA cocktail, rather than assigning a finite classification. Commonly, active learning strategies label samples where the model is most uncertain; however, this requires access to the posterior probability, $\hat{p}(y|x)$, which is not known for regression problems. Secondly, our problem statement requires that we select a \textit{batch} of 40 CPA cocktails to be tested with each iteration. Many active learning approaches select only one optimal point to evaluate at a given iteration, but we require ``batch-aware'' strategies which consider what batch of points will \textit{together} provide our model the most unique and informative information at that given iteration. Sener and Savarese published the core-set approach after finding that traditional active learning methods did not work well for training CNNs in a batch setting \cite{senerActiveLearningConvolutional2018}. Similarly, Wu et al. detail several greedy sampling approaches for active learning on regression problems \cite{wuActiveLearningRegression2019}. Both of these approaches boil down to varying implementations of the k-center algorithm, which aims to select the point(s) that are farthest from the existing data. In this way, samples selected in the new batch $T_{C,i}$ will provide information on regions of the design space which are not currently represented in the training dataset $T_{K,i}$.

\begin{figure}[!htb]
    \centering
    \includegraphics[width=0.9\linewidth]{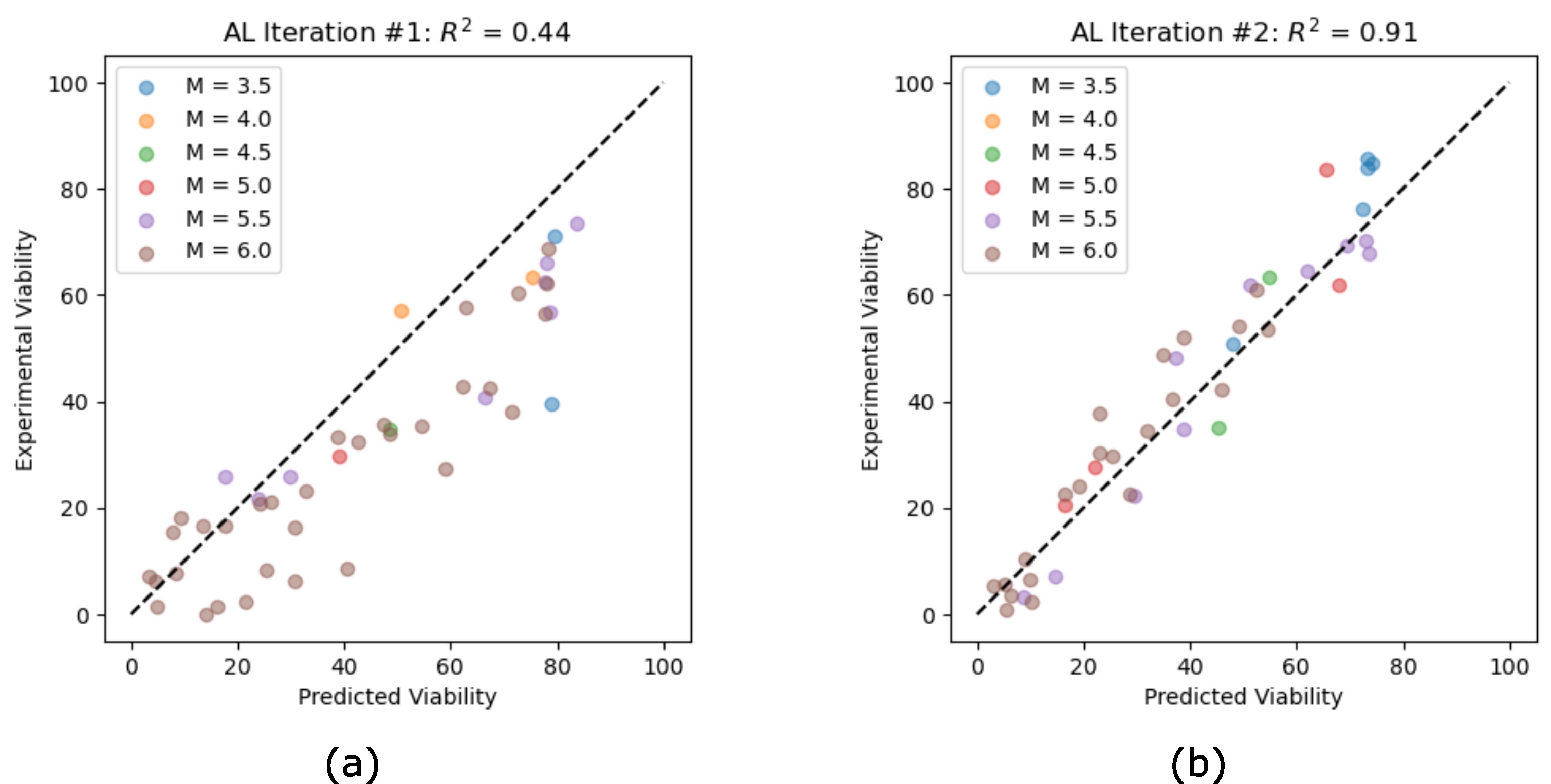}
    \caption{$R^2$ plots for the first two iterations of k-center active learning, with individual points colored according to total concentration of the cocktail. (a) For the first iteration, $R^2 = 0.44$, while (b) the second iteration improved to $R^2 = 0.91$.}
    \label{fig:active-learning-r2}
\end{figure}

We performed two iterations of the k-center active learning algorithm and plotted the coefficient of determination, $R^2$, for the model's initial viability prediction versus the experimentally determined viability in \cref{fig:active-learning-r2}. We use an XGBoost regression model as our surrogate model to predict cocktail viability \cite{chenXGBoostScalableTree2016}. We see significant improvement in the performance of the model after just one iteration of active learning, which is to be expected. Especially for the first iteration, the model has not seen any CPA cocktails with high 5.5 or 6.0M concentrations, and \cref{fig:active-learning-r2}a shows that the model overestimates the viability of almost all of the high concentration cocktails. In contrast, we can see from the $R^2$ plot of the second iteration in \cref{fig:active-learning-r2}b that the updated model much more accurately predicts the viability of the high concentration cocktails.

\section{Results}
We first demonstrate the efficacy of several Bayesian optimization methods on a synthetic model to build confidence in our implementation and optimize model hyperparameters. We then apply these methods with iterative high-throughput screening experiments to discover optimal CPA cocktails. We implement the ParEGO and EHVI acquisition functions into our Bayesian optimization framework using the BoTorch package \cite{balandatBoTorchFrameworkEfficient2020}. More specifically, we use the qLogNParEGO and qLogNEHVI acquisition functions. The `q' indicates that these acquisition functions can select batches of candidates, where $q$ is the batch size. In 2023, Ament et al. demonstrated that their proposed logarithmic versions of the expected improvement family of functions lead to significantly better performance, helping avoid the issue of vanishing gradients \cite{amentUnexpectedImprovementsExpected2025}, hence the `Log'. Lastly, the `N' indicates that these acquisition functions account for noise in the Gaussian process, which is important when considering the variance that can be present in our experimental data \cite{zhouCorrectedExpectedImprovement2023}.

\subsection{Experimental Bayesian Optimization}
In \cref{sec:appendix-b} we discuss how we optimized the hyperparameters for the various models discussed in this section using synthetic models in place of conduction real experiments. Once this is accomplished, we begin to experimentally evaluate CPA cocktails as suggested by our baseline and Bayesian optimization models. 

We begin with the initial set of 535 samples plus the two iterations of active learning, and perform 8 iterations of Bayesian optimization with a batch size of $q=10$ for each of the following methods: random sampling, qLogNParEGO, qLogNEHVI, qVarLogNEHVI. Here, the random sampling serves as the na\"ive method, qLogNParEGO as the baseline multi-objective optimization approach, and the two EHVI approaches as our state-of-the-art methods. The qVarLogNEHVI method is identical to the qLogNEHVI method introduced previously, except that it accounts for the variance seen in the experimental data. In theory, this allows the Gaussian process model to better capture uncertainty in the data.

When conducting the HTS experiments, we screen fourty experimental cocktails with each batch, or ten candidates for each of the four methods. The viabilities are averaged between $n = 3$ samples, each on different well plates. The viabilities are normalized using on-plate positive and negative controls. The turnaround time for an experimental batch is approximately one week. Importantly, each method is blind to data learned by the other methods. In this way, some methods may select the same candidates to evaluate, but we treat each sample independently.

\begin{figure}
    \begin{subfigure}[b]{0.45\linewidth}
        \centering
        \includegraphics[width=\linewidth]{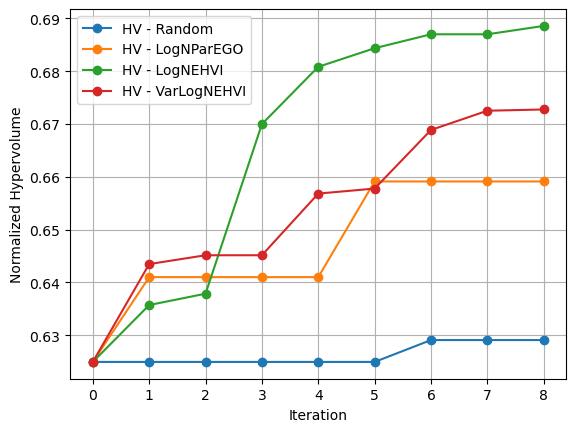}
        \caption{}
        \label{fig:exp_hv}
    \end{subfigure}
    \hfill
    \begin{subfigure}[b]{0.45\linewidth}
        \centering
        \includegraphics[width=\linewidth]{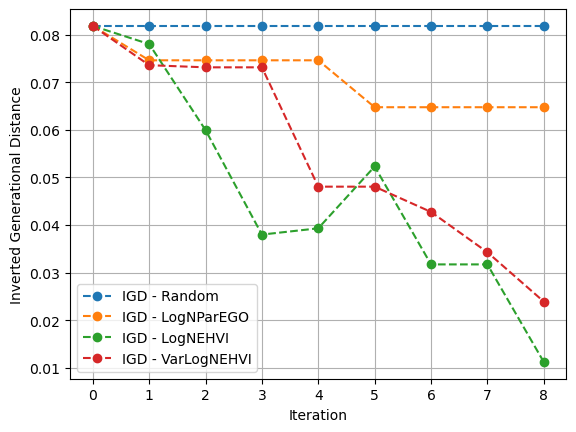}
        \caption{}
        \label{fig:exp_igd}
    \end{subfigure}
    \caption{Normalized Hypervolume (a) and Inverted Generational Distance (b) versus iteration for each of the four Bayesian optimization methods on the experimental data. Each method selects a batch of $q=10$ candidate CPA cocktails to evaluate at each iteration, with the experimental results being added to the training data for subsequent iterations.}
    \label{fig:exp_hv_igd}
\end{figure}

In \cref{fig:exp_hv_igd}, we plot two metrics to assess the multi-objective performance of our experiments. First, we plot the hypervolume, which measures the volume of the dominated portion of the objective space by the Pareto front. In our two objective case, this corresponds to the area of the Pareto front. A \textit{higher} hypervolume indicates a better Pareto front. Second, we plot the inverted generational distance (IGD), which measures the average distance from points on the true Pareto front to the nearest point on the estimated Pareto front \cite{ishibuchiComparisonHypervolumeIGD2019}. A \textit{lower} IGD indicates a closer approximation of the true Pareto front. Importantly, we normalize both of the objectives to the range $[0, 1]$ before calculating these metrics, such that no objective dominates the other.

With regard to hypervolume metric, we observe the two EHVI approaches significantly outperform the na\"ive and baseline methods, with the qLogNEHVI approach slightly outperforming the qVarLogNEHVI approach. The na\"ive approach, random sampling, only sees improvement to its hypervolume in the 6th iteration. The scalarization approach, qLogNParEGO, sees improvement in the 1st and 5th iterations, but otherwise plateaus. Both of the EHVI approaches see steady improvement over the iterations, with the qLogNEHVI approach performing the best overall.

With regard to the IGD metric, we observe similar trends. Recall that in the case of IGD, the closer to zero, the better. The na\"ive approach, random sampling, sees no improvement to its IGD overall all iterations. The scalarization approach, qLogNParEGO, sees improvement in the 1st and 5th iterations, but otherwise plateaus. he two EHVI approaches significantly outperform the na\"ive and baseline methods, with the qLogNEHVI approach slightly outperforming the qVarLogNEHVI approach. Interestingly, the LogNEHVI approach worsens between the 3rd, 4th and 5th iterations. This is possible due to the way the IGD is calculated. The formulation for IGD is detailed in \cref{eq:igd}. The IGD is measured as the average of the minimum distances from points on the true Pareto front, $P$, to the nearest point on the estimated Pareto front, $P^*$. Since we do not know the true Pareto front, we take the true Pareto front to be the union of all sampled points across the four methods at the end of the eight iterations, as well as the initial 535 points. In this way, the IGD cannot be calculated over the course of the iterations as the hypervolume can. As the set of points in the estimated Pareto front changes, it is possible for the average distance to increase even if new dominating Pareto optimal points are discovered. Regardless, we see consistent performance from the LogNEHVI and varLogNEHVI approaches overall.

\begin{equation}
    IGD(P, P^*) = \frac{1}{|P^*|} \sum_{y^* \in P^*} \left( \textrm{min}\left(\textrm{dist}(y^*, P)^2 \right)^{1/2} \right)
    \label{eq:igd}
\end{equation}

\begin{figure}[!htb]
    \centering
    \includegraphics[width=\linewidth]{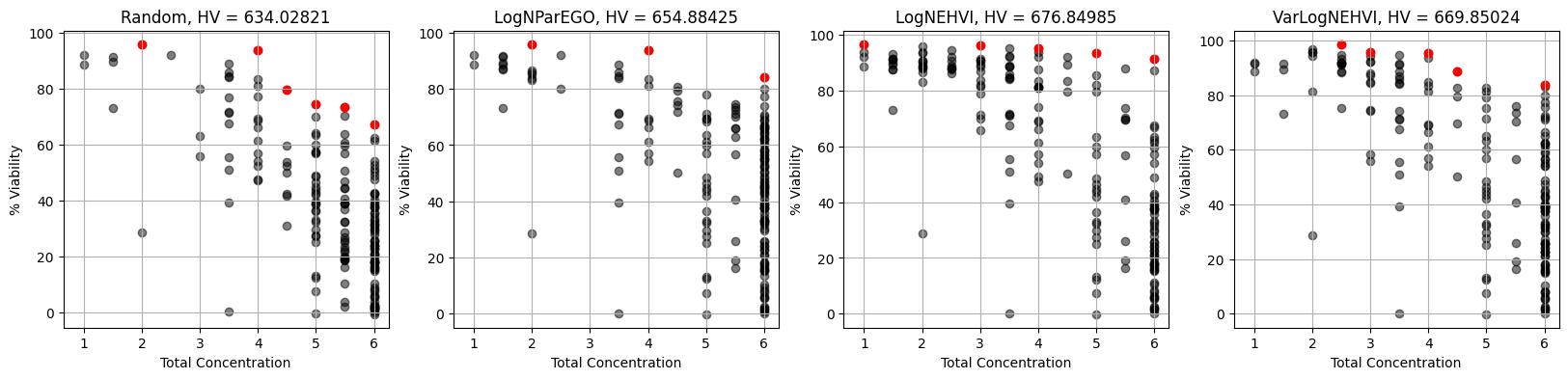}
    \caption{Visualization of the Pareto front for each of the four methods after 8 iterations of Bayesian optimization. The orange points represent the Pareto optimal points for the given method, while the gray points are in the dominated set.}
    \label{fig:exp_pareto}
\end{figure}

\begin{figure}[!htb]
    \centering
    \includegraphics[width=\textwidth]{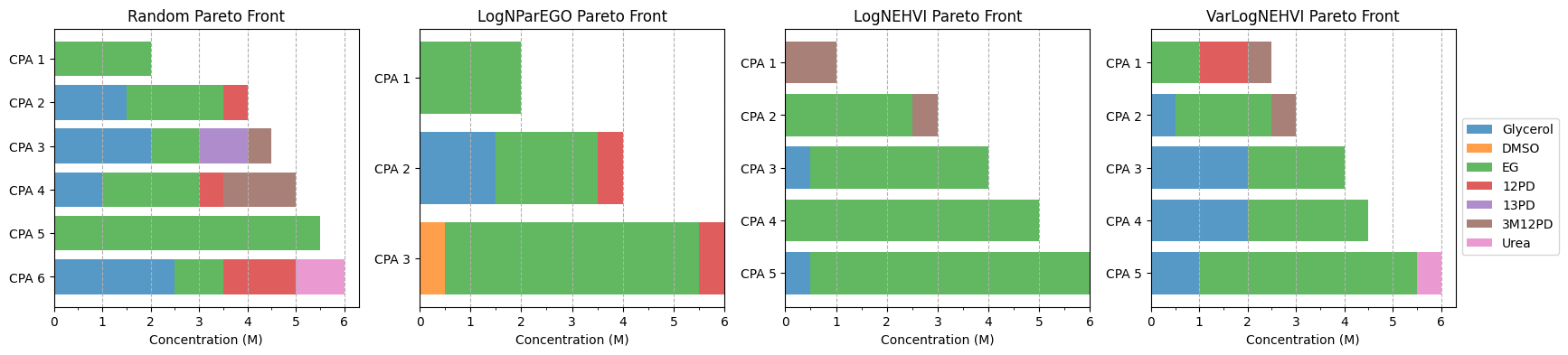}
    \caption{Composition of the experimentally determined Pareto optimal CPA cocktails for each of the four methods after 8 iterations of Bayesian optimization.}
    \label{fig:exp_composition}
\end{figure}

In \cref{fig:exp_pareto}, we plot the Pareto optimal points for each of the four methods after all 8 iterations. This allows us to visualize the objective space between the four methods. We observe that the best performing method, qLogNEHVI, has discovered several high viability cocktails with high concentrations in the 4-6M range. In \cref{fig:exp_composition}, we plot the composition of the Pareto optimal cocktails for each of the four methods after all 8 iterations. Here we can visualize the composition of the Pareto optimal points from \cref{fig:exp_pareto}. We observe that all three non-naive methods have a strong preference for EG-rich cocktails, which makes sense because the T24 assay is particularly resistant to EG toxicity. The naive method samples a much more diverse set of cocktails due to its random nature, but we can observe in \cref{fig:exp_pareto} that, especially in the high concentration range, these cocktails do not yield high viabilities.

\section{Discussion}

Overall, this study demonstrates the effectiveness of Bayesian optimization for discovering high-performing CPA cocktails in both synthetic benchmarks and iterative experimental settings. In particular, hypervolume-based optimization strategies consistently outperformed naive random sampling and strong scalarization-based baselines, yielding higher-quality Pareto fronts under constrained experimental budgets. In the present study, experimental exploration was limited to a maximum total CPA concentration of 6~M and discrete 0.5~M concentration increments for individual components, reflecting practical constraints of the high-throughput screening (HTS) platform. With an expanded design space and a more diverse initial dataset, hypervolume-based methods are expected to uncover an even broader and potentially more informative set of optimal formulations. Importantly, these constraints arise from assay and platform limitations rather than from the Bayesian optimization methodology itself.

Accurate assessment of CPA toxicity at high concentrations remains experimentally challenging, particularly due to confounding effects from osmotic shrinkage and swelling that can obscure true cytotoxic responses \cite{mazurOsmoticResponsesPreimplantation1986,jaskiewiczValidationHighthroughputScreening2025,kangasEliminatingOsmoticStress2025}. In addition, measured viability is sensitive to experimental conditions such as temperature, exposure duration, and the timing and sequence of CPA loading steps \cite{mazurTwofactorHypothesisFreezing1972}. Ensuring strict control and repeatability of these factors is therefore essential for reliable data generation. In this study, viability measurements were obtained using a limited number of replicates ($n = 3$), which can introduce substantial variance and sensitivity to outliers. In this context, the qVarLogNEHVI approach is particularly advantageous, as it explicitly accounts for experimental variability when modeling uncertainty. By incorporating observation noise into the Gaussian process surrogate, this method provides more faithful uncertainty estimates, leading to improved predictive accuracy and more robust selection of candidate CPA cocktails for subsequent evaluation.

At a broader level, this study demonstrates the effectiveness of integrating high-throughput experimental screening with machine learning and Bayesian optimization to efficiently explore and optimize a complex formulation design space. By enabling targeted experiment selection, the proposed framework facilitates the identification of synergistic interactions among CPA components that are difficult to infer using conventional experimental strategies or mechanistic reasoning alone. Notably, the optimization-driven selection of candidate cocktails frequently identified high-performing formulations that would not have been prioritized based on expert intuition, reinforcing the value of data-driven decision-making in this context. Because the framework relies only on the availability of a suitable assay and a defined formulation space, it is readily extensible to alternative cryoprotectant libraries, cell lines, and experimental conditions.

\section{Conclusion and Future Directions}

In this work, we presented a data-efficient framework for CPA cocktail design that combines high-throughput experimental screening with multi-objective Bayesian optimization. The approach targets the central vitrification trade-off between achieving ice-suppressing CPA concentrations and maintaining post-exposure cell viability. By formulating cocktail discovery as a black-box, multi-objective optimization problem, the framework enables systematic exploration of a large formulation space that would be impractical to exhaustively evaluate using conventional experimental methods.

Our experimental results show that hypervolume-based Bayesian optimization methods, particularly expected hypervolume improvement variants, consistently outperform random sampling and strong scalarization-based baselines. Across iterative wet-lab experiments, these methods more rapidly improve Pareto front quality, as measured by hypervolume and inverted generational distance, while requiring substantially fewer experimental evaluations. Complementary synthetic studies further confirm the sample efficiency of the approach, demonstrating recovery of comparable Pareto-optimal solution sets using a fraction of the evaluations required by prior state-of-the-art methods.

Beyond performance gains, this work provides several broader insights for the cryobiology and biomedical engineering communities. We argue that uncertainty-aware surrogate modeling is critical when working with noisy biological assays and limited data, enabling principled balancing of exploration and exploitation. Framing CPA cocktail discovery as a multi-objective problem more naturally captures biological trade-offs and avoids ad hoc scalarization choices. Additionally, integrating experimental automation with machine learning can reveal synergistic CPA interactions that may not be apparent from expert intuition alone.

Several avenues for future work remain. Expanding the formulation space to finer concentration resolutions, additional cryoprotectants, or non-penetrating additives may further enhance cocktail performance. Incorporating additional objectives such as osmotic stress or post-thaw functional outcomes would support more application-specific optimization. Methodologically, adaptive refinement of surrogate models and acquisition strategies could improve robustness as data accumulate, while application to different cell lines, tissues, or organ-scale systems represents a key step toward broadly generalizable cryopreservation design.

\newpage
\printbibliography
\newpage
\section{Appendix A: Bayesian Optimization Background}
\label{sec:appendix-a}

\subsection{Exploration-Exploitation Dilemma}
\begin{figure}[!htb]
    \centering
    \includegraphics[width=0.8\linewidth]{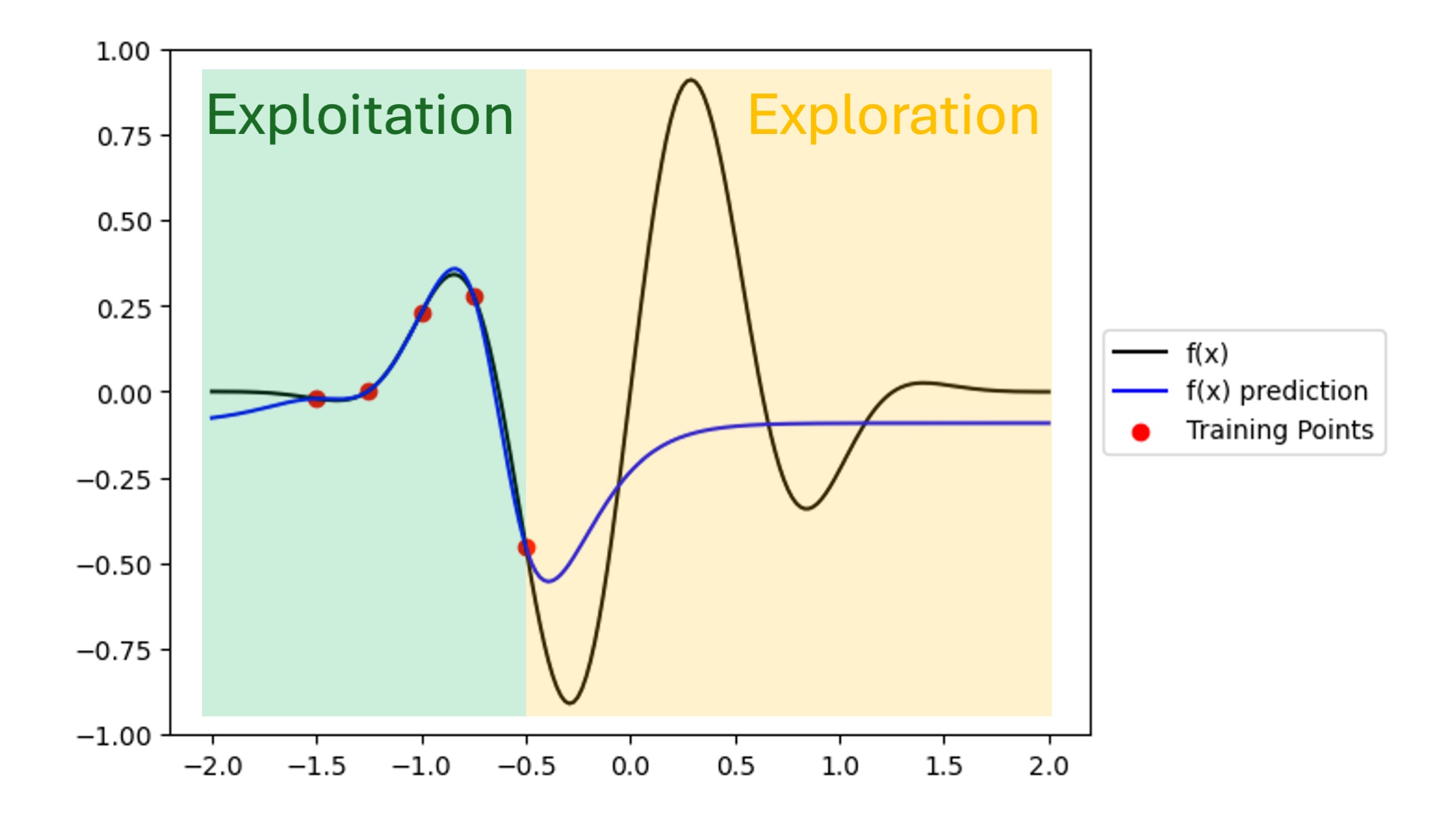}
    \caption{Simple function demonstrating the exploration-exploitation dilemma. The blue line represents our model's current representation of the true unknown function based on the red training points. The true function is plotted as the black line.}
    \label{fig:exploration-exploitation}
\end{figure}

We saw how quickly active learning improved our machine model's prediction in \cref{sec:active-learning}. However, while this active learning approach helps improve our model's coverage of the design space, it does not directly help us discover optimal solutions. We could simply sample the maximum points according to our current model, but this presumes that our model accurately captures the behavior of the underlying toxicity function at all points in the design space. This brings us to a concept in machine learning known as the ``exploration-exploitation dilemma''. When selecting candidate points to evaluate, we must balance between exploitation and exploration. Exploitation considers selecting the optimal points based on our model's current state. Exploration considers selecting unique points about which we have little certainty, but which may lead to better outcomes in the future. The previously discussed k-center active learning approach corresponds to a purely explorative approach. \cref{fig:exploration-exploitation} plots a simple function to demonstrate the exploration-exploitation dilemma. Here we have information on five training points on the right-hand side of the domain. If we used a purely exploitative approach, we would incorrectly identify the local maximum as the maximum of the domain. Instead, we require exploration of the right-hand side of the domain, such that when we utilize the exploitative approach, we can correctly identify the true maximum of the function. Therefore, we require a method that balances both exploration and exploitation. 

\subsection{Bayesian Optimization}
Perhaps one of the best-known methods for balancing exploration and exploitation of a black box function is Bayesian optimization. Bayesian optimization has two primary components: a probabilistic surrogate model, typically a Gaussian process regression model, and an acquisition function, which determines the next most optimal point(s) to sample \cite{shahriariTakingHumanOut2016}. Right away, we observe that Bayesian optimization has the two components our CPA cocktail optimization pipeline requires: a machine learning model that attempts to model the underlying CPA cocktail toxicity mechanisms, and a method to determine the next best points to sample. 

\subsubsection{Gaussian Process Regression}
Like any other machine learning regression model, the Gaussian process model returns some estimate of the mean behavior of our function, $\mu(\mathbf{x})$, for any combination of inputs in the design space. Gaussian processes are unique in the sense that they also provide an uncertainty, $\sigma(\mathbf{x})$, around each of these points. This distinction is incredibly important when we consider that we would like to sample points where our model believes there to be an optimal value, or where our model has a high degree of uncertainty. It is this important characteristic of the Gaussian process model that allows us to consider both.

\subsubsection{Acquisition Functions}
Now that we have our regression model selected, we need a method to determine the next best points to sample. While there are many possible acquisition functions, they all serve the same goal: selecting the next point(s) to sample by balancing both exploration and exploitation. This is perhaps most easily seen in the upper confidence bound acquisition function detailed in \cref{eq:ucb}, where the parameter $\beta$ can be seen as a weighting between exploitation and exploration. For low values of $\beta$, the acquisition function is highly exploitative, prioritizing where the mean $\mu(\mathbf{x})$ of the Gaussian process is at a maximum. For high values of $\beta$, the acquisition function is highly explorative, prioritizing where the Gaussian process is highly uncertain, with high values of $\sigma(\mathbf{x})$. Ideally, one would find a value of $\beta$ which balances this trade-off between exploration and exploitation, but this parameter must be tuned to the specific problem.

\begin{equation}
    \label{eq:ucb}
    a_{UCB}(\mathbf{x}) = \mu(\mathbf{x}) + \beta \cdot \sigma(\mathbf{x})
\end{equation}

Perhaps the most common acquisition function is the expected improvement acquisition function, detailed in \cref{eq:ei}. The expected improvement does not have any parameters that need to be set, and instead explicitly balances the exploration and exploitation trade-off. The first term correlates to exploitation, that is to say, when the mean value of the Gaussian process, $\mu(\mathbf{x})$, is higher than the previously highest function value, $f'$, the acquisition function will take a high value. The second term correlates to exploration, when the Gaussian process is highly uncertain and therefore has a high value of $\sigma(\mathbf{x})$, the acquisition function will take a high value.

\begin{equation}
    \label{eq:ei}
    a_{EI}(\mathbf{x}) = (\mu(\mathbf{x}) - f')\Phi\left(\frac{\mu(\mathbf{x}) - f'}{\sigma(\mathbf{x})}\right) + \sigma(\mathbf{x})\phi\left(\frac{\mu(\mathbf{x}) - f'}{\sigma(\mathbf{x})} \right)
\end{equation}

There are numerous other acquisition functions, all with their own strengths and weaknesses. It has been observed that the optimal acquisition function strategy can change over the course of a given optimization process \cite{shahriariTakingHumanOut2016}. Therefore, when implementing our optimization pipeline, it is critical to survey a range of acquisition functions and hyperparameters (if present).

\subsubsection{Bayesian Optimization Example}
In this section, we will present several examples of Bayesian optimization on one and two-dimensional synthetic functions to illustrate its utility. The objective function, which we are trying to optimize in the one-dimensional case, is presented in \cref{eq:bo-1d-func} over the range $x \in [-2, 2]$. The maximum in this range is approximately $f(0.289) = 0.909$. We note that this is the same function used to illustrate the exploration-exploitation trade-off in \cref{fig:exploration-exploitation}.

\begin{equation}
    \label{eq:bo-1d-func}
    f(x) = \text{sin}(5x) \cdot \left(1 - \text{tanh}(x^2) \right)
\end{equation}

\begin{figure}[!htb]
    \centering
    \includegraphics[width = \linewidth]{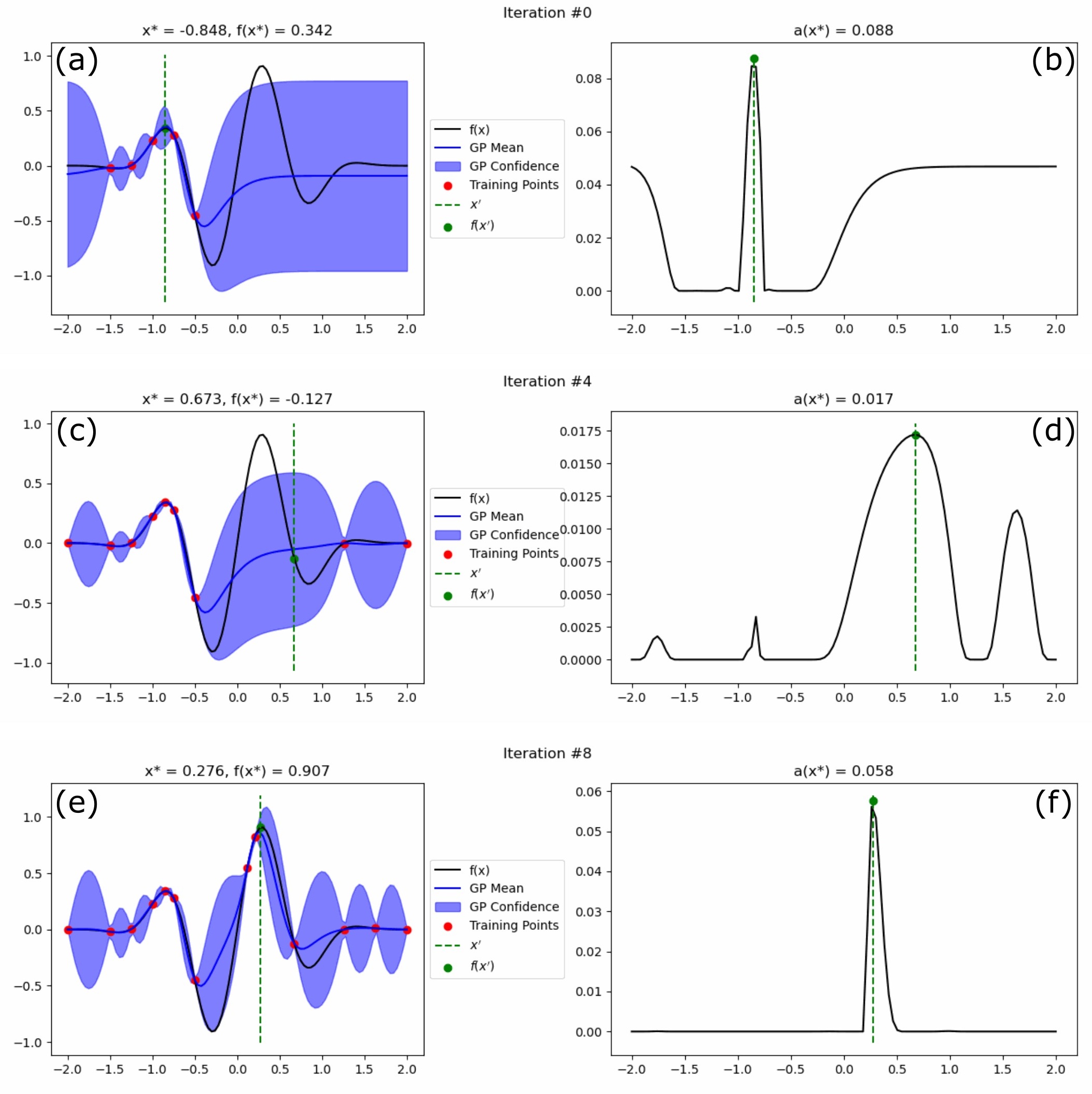}
    \caption{Bayesian optimization iterations for the one-dimensional function presented in \cref{eq:bo-1d-func}, using the expected improvement acquisition function. The true function, current Gaussian process mean and confidence interval, training points, and currently selected point is visualized on the objective function plots in (a), (c), and (e), while the acquisition function and currently selected point is visualized in the acquisition function plots in (b), (d), and (f), for their respective iterations.}
    \label{fig:bo-1d}
\end{figure}

To illustrate the importance of exploration and exploitation, we initialize the model with five training points, all on the left-hand side of the domain. If the acquisition function does not properly balance exploration and exploitation, it could end up getting stuck on the local minima on the left-hand side. We use the expected improvement acquisition function and are attempting to maximize the function. We observe in \cref{fig:bo-1d} (a) and (b), the acquisition function targets the local maximum. Once information on this maximum has been obtained, we see that the next couple of iterations begin to target the unexplored right-hand side of the domain, as seen by the newly added training points and currently selected point in (c) and (d). In this way, the expected improvement is balancing both exploration and exploitation. Finally, in the eighth iteration, in (e) and (f), we see that we have converged quite close to the true maximum of the function, with our current guess of $f(0.276) = 0.907$. We can observe that in (f), the acquisition function is fairly certain that the maximum worth sampling lies exactly at that point, whereas in (d), there are several regions where the acquisition function has peaks. Furthermore, we can observe in (e) that as more training points are added to the domain, the confidence intervals on the Gaussian process model begin to shrink.

Next, we present the application of batch Bayesian optimization on a two-dimensional synthetic function with many local minima. The synthetic function is known as the Rastrigin function, as detailed in \cref{eq:rastrigin}, where $A = 10$, and $n$ is the dimension of $\mathbf{x}$, in this case, 2. We consider values in the range $\mathbf{x} \in [-2.5, 2.5]$, and we are attempting to minimize this function. The true minimum exists at $f(0,0) = 0$.

\begin{equation}
    \label{eq:rastrigin}
    f(\mathbf{x}) = A\cdot n + \sum_{i=1}^n \left[x_i^2 + A \text{cos}(2\pi x_i) \right]
\end{equation}

\begin{figure}[!htb]
    \centering
    \includegraphics[width=0.9\linewidth]{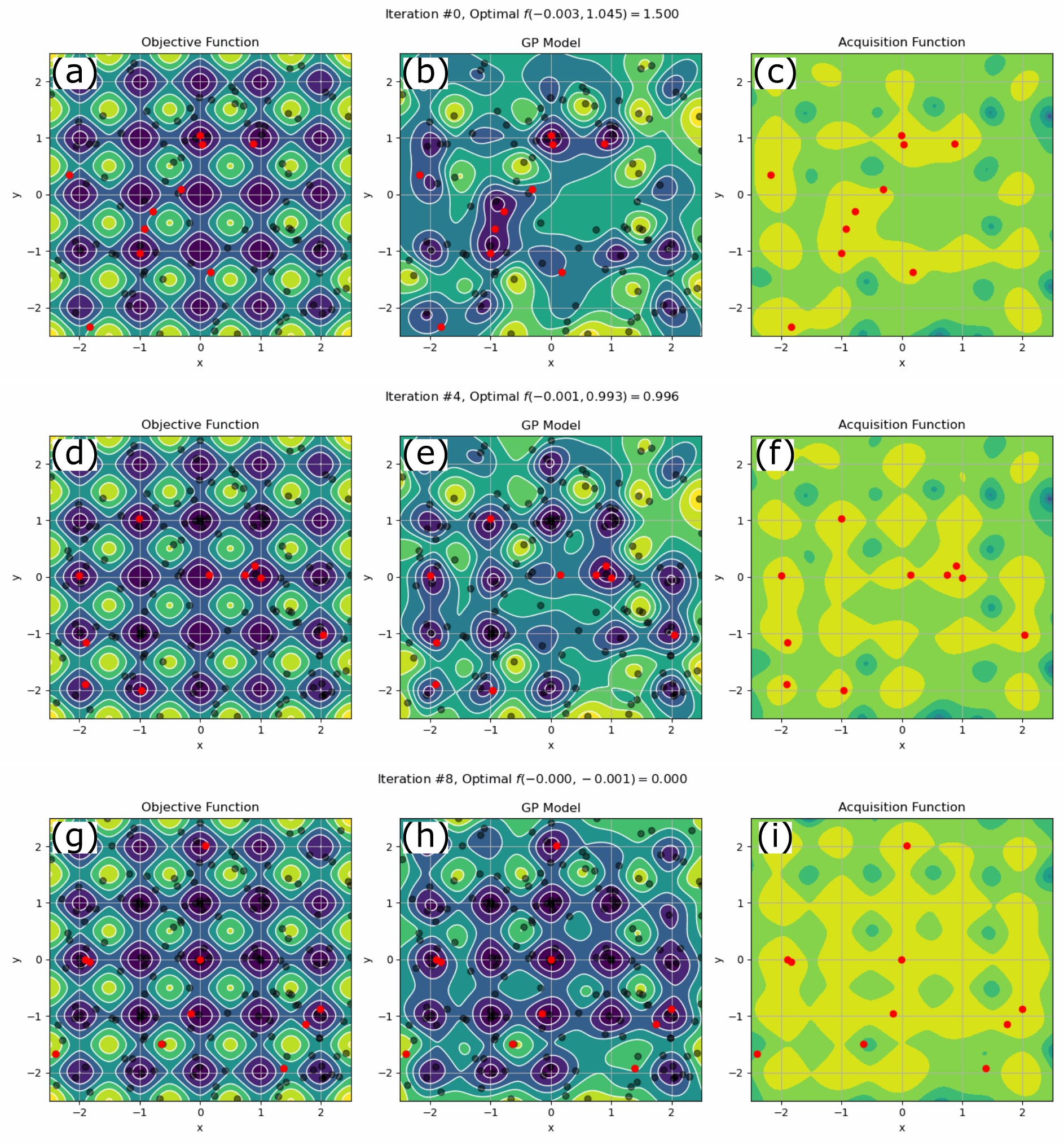}
    \caption{Bayesian optimization iterations for the two-dimensional Rastrigin function presented in \cref{eq:rastrigin}, using the batch expected improvement acquisition function. The true function, training points (gray), and candidate points (red) are visualized in the contour plots in (a), (d), and (g). The Gaussian process regression model's current representation of the function is visualized in the contour plots in (b), (e), and (h), along with the training points and candidate points. Note: We cannot visualize the uncertainty of the GP on the same plot since this would require an additional dimension, but the uncertainty of the GP model is still considered when determining the value of the acquisition function. Finally, the acquisition function and the selected candidate points are visualized in the contour plots in (c), (f), and (i).}
    \label{fig:bo-2d}
\end{figure}

Again, we use the expected improvement acquisition function. However, since there are many local minima in the domain, we consider a batch application of the expected improvement. With each iteration, we select 10 candidate points, also known as a batch size of $q=10$. To evaluate the batch version of the expected improvement acquisition function, we must use Monte Carlo sampling since there is no closed-form expression for the function \cite{balandatBoTorchFrameworkEfficient2020,amentUnexpectedImprovementsExpected2025}. Therefore, when implementing batch Bayesian optimization, there are additional hyperparameters to set with regard to the Monte Carlo sampling strategy that we will not delve into here. When developing a pipeline, it is important that these hyperparameters are optimized. We visualize several iterations of the batch Bayesian optimization in \cref{fig:bo-2d}. We observe significant improvement in the accuracy of the GP model fit from the zeroth iteration in (b) to the eighth iteration in (h). Furthermore, the true minimum of the objective function has been nearly found by the eighth iteration, as shown in (g), (h), and (i). We observe in (c), (f), and (i) that the batch sampling strategy effectively samples points from various maxima across the acquisition function domain within each iteration, effectively balancing exploration and exploitation. 

\subsection{Multi-Objective Optimization}
\begin{figure}[!htb]
    \centering
    \begin{subfigure}[b]{0.45\linewidth}
        \centering
        \includegraphics[width=\linewidth]{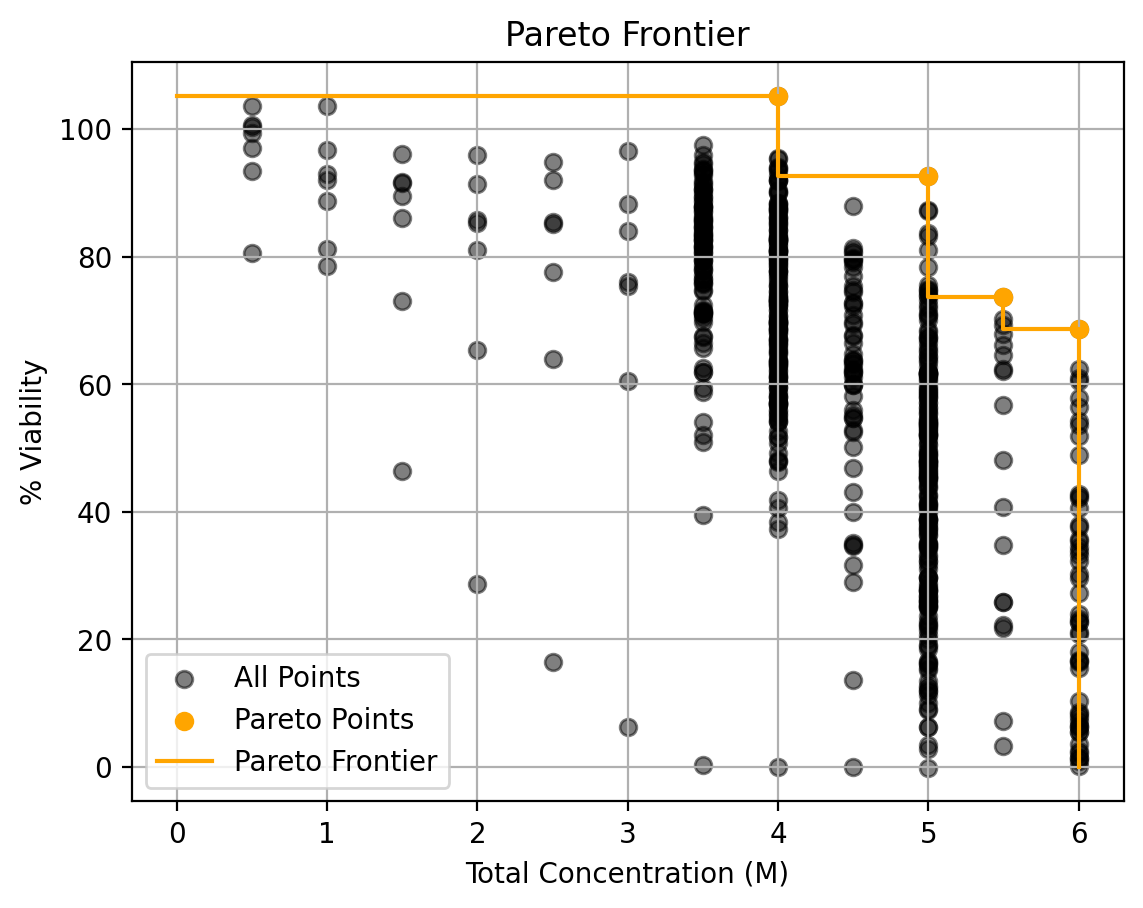}
        \caption{}
        \label{fig:pareto}
    \end{subfigure}
    \hfill
    \begin{subfigure}[b]{0.45\linewidth}
        \centering
        \includegraphics[width=\linewidth]{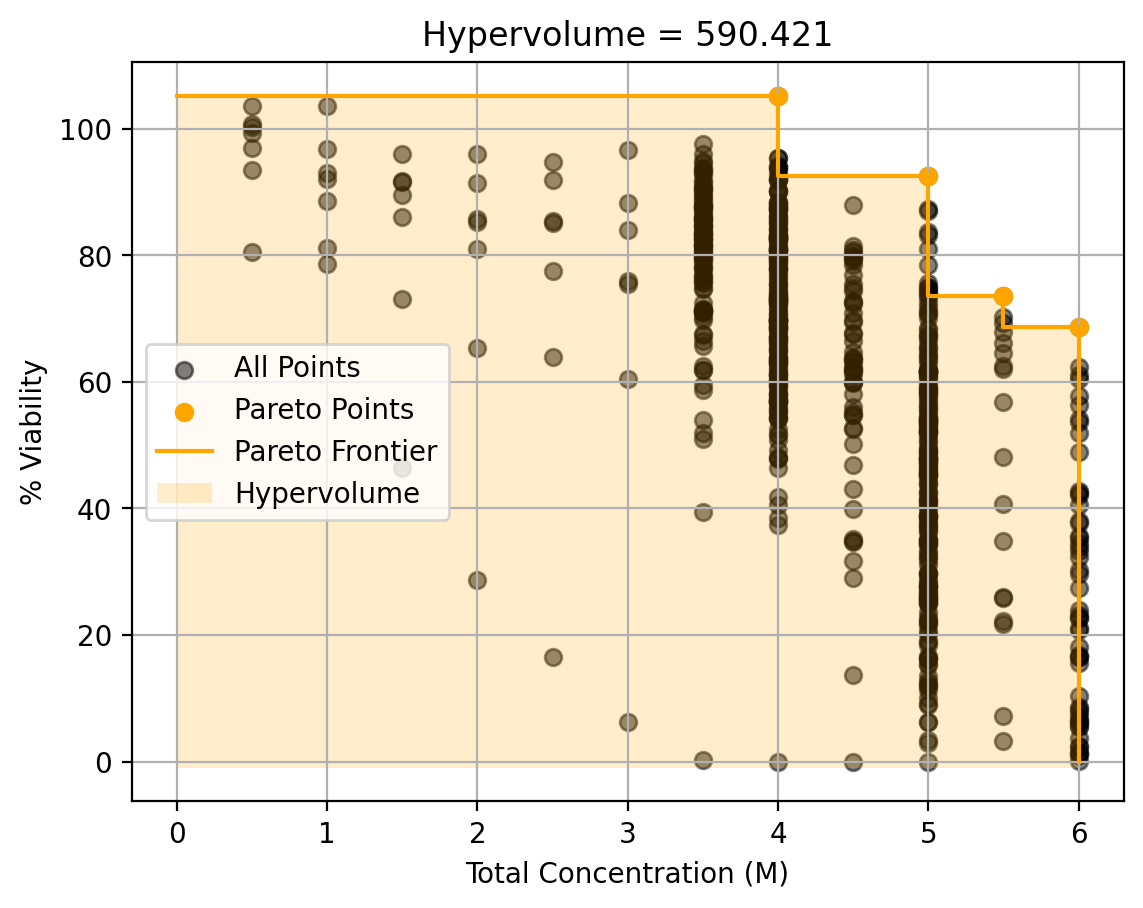}
        \caption{}
        \label{fig:hypervolume}
    \end{subfigure}
    \caption{Visualization of the Pareto front \ref{fig:pareto} and hypervolume \ref{fig:hypervolume} for our initial training data.}
\end{figure}

The previous cases consider Bayesian optimization, where we are trying to maximize or minimize a single objective function. However, in our application, we are simultaneously aiming to discover CPA cocktails with both high concentration and high viability. The goal in multi-objective Bayesian optimization is to obtain the Pareto front, which contains the set of all optimal CPA cocktails when considering both concentration and viability. The Pareto front represents a set of trade-offs in the objective space, namely, optimizing one objective leads to a deterioration of the other \cite{daultonParallelBayesianOptimization2021}. In our case, we would expect that as the concentration of the CPA cocktail increases, its viability will decrease, and vice versa. The Pareto front for our initial dataset is visualized in \cref{fig:pareto}. The orange points represent the set of Pareto optimal points, while all other points are dominated by at least one point in the Pareto front. A point is said to dominate another point if it is better in at least one objective and not worse in any other objective.

There are several approaches to multi-objective Bayesian optimization, but we will focus on two of the most common: scalarization and hypervolume approaches. Scalarization methods convert the multi-objective problem into a single-objective problem by combining the objectives into a single scalar value. Hypervolume methods consider the volume of the objective space that is dominated by the Pareto front, and aim to maximize this volume. The hypervolume for our initial dataset is visualized in \cref{fig:hypervolume}, where the reference point is the origin $\left(0,0\right)$.

\subsubsection{Scalarization Methods}
Scalarization methods convert the multi-objective problem into a single-objective problem by combining the objectives into a single scalar value. Perhaps the simplest scalarization method is the weighted sum approach, where each objective is assigned a weight, and the objectives are combined into a single scalar value. The weighted sum approach is given by \cref{eq:weighted-sum}, where $w_i$ is the weight for objective $i$, and $f_i(\mathbf{x})$ is the value of objective $i$ for input $\mathbf{x}$.

\begin{equation}
    f(\mathbf{x}) = \sum_{i=1}^n w_i f_i(\mathbf{x})
    \label{eq:weighted-sum}
\end{equation}

While the weighted sum approach is simple to implement and understand, it has several limitations. First, it requires the user to specify the weights for each objective, which can be difficult to do in practice. Secondly, the weighted sum approach is only optimal for convex objective functions. This means if the Pareto front is non-convex, the weighted sum approach may not be able to find all Pareto optimal points \cite{marlerWeightedSumMethod2010}.

Instead, we utilize the Pareto Efficient Global Optimization (ParEGO) approach, first introduced by Knowles in 2006 \cite{knowlesParEGOHybridAlgorithm2006}. ParEGO utilizes augmented Chebyshev scalarization, where scalarization weights are randomly sampled from a uniform distribution, providing good coverage of the objective space as the weights vary. In this sense, ParEGO still converts our multi-objective problem into a single-objective problem, and we can still apply the classical Bayesian optimization techniques.

\subsubsection{Hypervolume Methods}
\begin{figure}[!htb]
    \centering
    \includegraphics[width=0.5\linewidth]{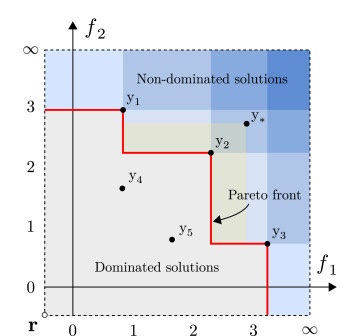}
    \caption{Hypervolume improvement visualized for a two-objective problem. The yellow shaded region represents the hypervolume improvement associated with the new candidate's point $\mathbf{y}_*$. Figure adapted from \cite{kansaraMultiobjectiveBayesianOptimisation2025}.}
    \label{fig:hv-improvement}
\end{figure}

More recently, hypervolume-based methods have been developed for multi-objective Bayesian optimization. These methods simultaneously consider all objectives by considering the hypervolume of the Pareto-dominated space. This simply requires a reference point that is worse than all existing points, and then the hypervolume can be computed in any dimension. For a two-objective problem, this corresponds to the area of the dominated space, whereas for a three-objective problem it would be the volume of the dominated space, and so on. The notion of hypervolume provides us with a useful metric to quantify the overall quality of our set of Pareto optimal points. Furthermore, we can consider the hypervolume improvement, which is the increase in hypervolume from a new candidate point. We visualize the hypervolume improvement in the case of a two-objective problem in \cref{fig:hv-improvement}. Here we see that the new candidate point $\mathbf{y}_*$ provides a significant increase in hypervolume, as represented by the yellow-shaded region. In contrast, the points $y_4$ and $y_5$ do not provide any hypervolume improvement, as they are dominated by the existing Pareto front.

The concept of hypervolume improvement can be used to define an acquisition function similar to the expected improvement acquisition function that we introduced in \cref{eq:ei}. The expected hypervolume improvement (EHVI) acquisition function considers the expectation of the hypervolume improvement for a given candidate point based on the current Gaussian process models.
\section{Appendix B: Multi-Objective Bayesian Optimization Model Development}
\label{sec:appendix-b}

\subsection{Synthetic Model Bayesian Optimization}
\begin{figure}[!htb]
    \centering
    \begin{subfigure}[b]{0.54\linewidth}
        \centering
        \includegraphics[width=\linewidth]{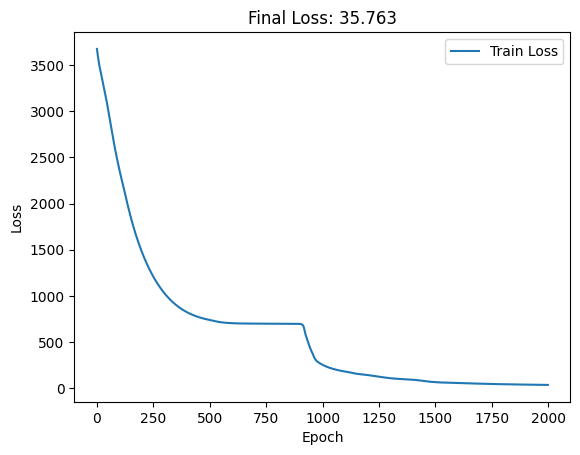}
        \caption{}
        \label{fig:nn_loss}
    \end{subfigure}
    \hfill
    \begin{subfigure}[b]{0.42\linewidth}
        \centering
        \includegraphics[width=\linewidth]{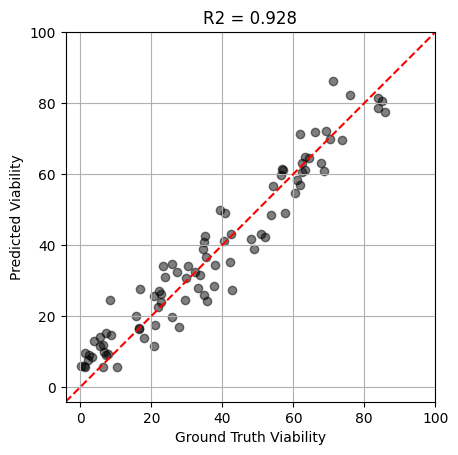}
        \caption{}
        \label{fig:nn_r2}
    \end{subfigure}
    \caption{Neural network surrogate model trained to predict viability provided composition of CPA cocktail. \ref{fig:nn_loss} depicts training loss vs. epoch, and \ref{fig:nn_r2} depicts $R^2$ on the candidates selected during the first two iterations of active learning.}
\end{figure}

Before running real experiments on unseen CPA cocktails, we use the initial set of 535 samples to perform synthetic experiments. This allows us to optimize hyperparameters for each of the approaches, as well as build confidence that our chosen approaches work in practice. In place of the HTS experiments, we train a neural network to predict the viability of CPA cocktails based on the composition of the cocktail. In their seminal paper, Hornik et al. prove that MLPs are universal function approximators, meaning that our neural network should be able to capture the underlying mechanisms of CPA toxicity, provided the right data \cite{hornikMultilayerFeedforwardNetworks1989}. The neural network is a multilayer perceptron (MLP) with three hidden layers of size [16, 32, 16], and sigmoid activation functions on all layers except the output. We use the Adam optimizer with a learning rate of 0.01 and train for 2000 epochs. The model is trained on the initial set of 535 samples, which we partition into training and test sets using k-fold cross-validation with $k=5$. We use the mean squared error (MSE) as our loss function, and we evaluate the model's performance using the MSE loss as well as the coefficient of determination, $R^2$, on the test set. In \cref{fig:nn_loss}, we visualize the MSE loss during training of our neural network. In \cref{fig:nn_r2} we visualize the $R^2$ plot of the model's predictions on unseen test points, specifically the points selected during the first two iterations of active learning. We observe a high $R^2$ value of 0.928 on this unseen test set, building confidence that our model is well-trained.

\begin{figure}[!htb]
    \centering
    \begin{subfigure}[b]{0.48\linewidth}
        \centering
        \includegraphics[width=\linewidth]{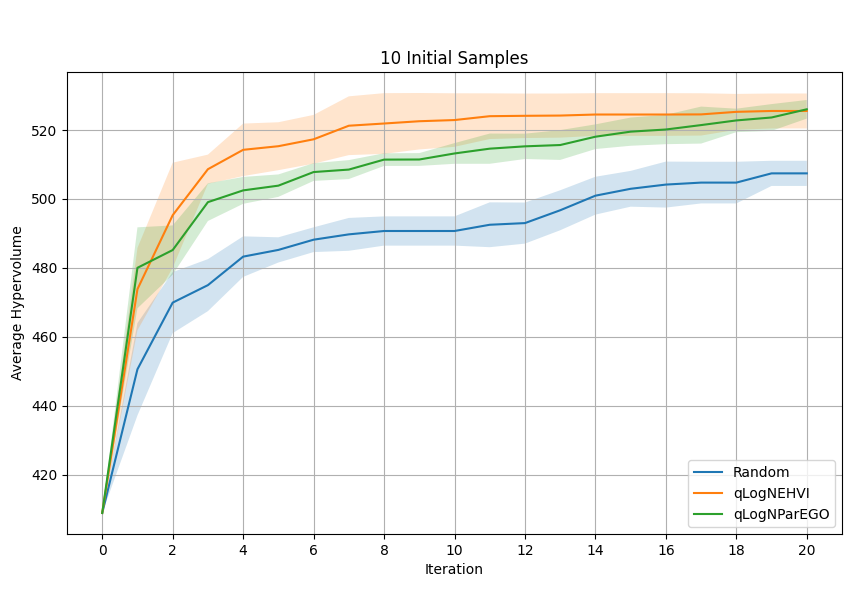}
        \caption{}
        \label{fig:syn_hv_k10}
    \end{subfigure}
    \hfill
    \begin{subfigure}[b]{0.48\linewidth}
        \centering
        \includegraphics[width=\linewidth]{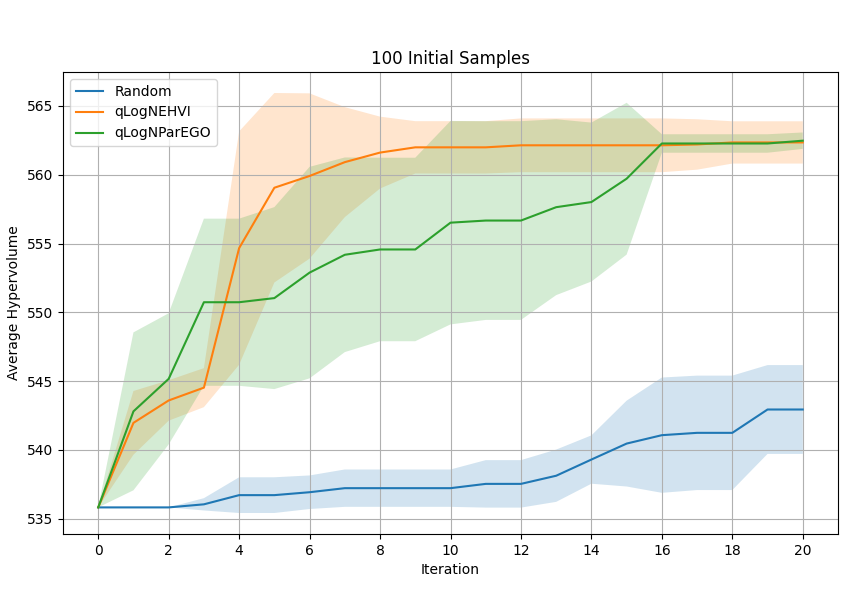}
        \caption{}
        \label{fig:syn_hv_k100}
    \end{subfigure}
    \caption{Hypervolume versus iteration for various Bayesian optimization methods on the synthetic model. \ref{fig:syn_hv_k10} is initialed with $k=10$ samples, while \ref{fig:syn_hv_k100} is initialized with $k=100$ samples. The optimizations are run 10 times each, for 20 iterations, with a batch size of $q=10$. The shaded regions represent one standard deviation from the mean hypervolume across the 10 runs.}
\end{figure}

Now that we have a functioning surrogate model to query in place of conducting real experiments, we can run iterations of Bayesian optimization. We observe that the initial set of samples provides a very strong coverage of the Pareto optimal space under the surrogate model. Specifically, the initial 535 samples hypervolume of 511.1. If we query the surrogate model for all 50,000 possible CPA cocktails with concentrations under 6M, we find the maximum hypervolume to be 520.5. We determined that this was not a sufficiently challenging optimization problem for the synthetic case, so we reduced the initial dataset to only 10 or 100 random samples from the original set of 535 samples and the first two iterations of active learning. Using k-center clustering, we selected as diverse a set of samples as possible. We then performed 20 iterations of Bayesian optimization with a batch size of $q=10$ for each of the following methods: random sampling, qLogNParEGO, and qLogNEHVI. We ran each optimization 10 times to account for the stochasticity in the methods. We plot the hypervolume versus iteration for each of the three methods in \cref{fig:syn_hv_k10} and \cref{fig:syn_hv_k100}. The solid line corresponds to the mean hypervolume across the 10 runs, while the shaded region corresponds to one standard deviation from the mean. We observe in the $k = 10$ case that there is significant room for improvement in all three methods, especially in early iterations, with an initial hypervolume of ~400. That said, the two Bayesian optimization methods significantly outperform random sampling, with qLogNEHVI slightly outperforming qLogNParEGO. In the $k = 100$ case, we observe that the initial hypervolume is already quite high at 535, so there is not as much room for improvement. Still, we see that both Bayesian optimization methods perform quite well, while the random sampling method struggles to improve on the already rich initial dataset. In \cref{fig:syn_hv_k100}, we observe that the qLogNEHVI approach converges to a maximum hypervolume of ~562 after about 10 iterations, while the qLogNParEGO approach takes about 16 iterations to reach this point. In this sense, we see that the qLogNEHVI approach is more sample-efficient than the qLogNParEGO approach.

Ideally, we would like our Bayesian optimization approach to discover the Pareto optimal set of CPAs as efficiently as possible. Through these synthetic experiments, we demonstrated that Bayesian optimization approaches are sample-efficient and able to discover more optimal Pareto fronts than naive methods like random sampling. Furthermore, we observed that the qLogNEHVI approach was more sample efficient than the qLogNParEGO approach. Therefore, we are confident in our implementation of these methods and can now apply them to real experiments.

\end{document}